\documentclass[final,5p,times,twocolumn]{elsarticle}

\usepackage{amssymb}

\usepackage{adjustbox}
\usepackage{booktabs}
\usepackage{blindtext}
\usepackage{verbatim}
\usepackage{xspace}
\usepackage{newtxtext,newtxmath}
\usepackage{graphicx}
\journal{Knowledge-Based Systems}

\begin{document}

\begin{frontmatter}



\title{HiTSKT: A Hierarchical Transformer Model for Session-Aware Knowledge Tracing}


\author[inst1]{Fucai Ke}
\author[inst1]{Weiqing Wang}
\author[inst1]{Weicong Tan}
\author[inst1]{Lan Du}
\author[inst1]{Yuan Jin}
\author[inst1]{Yujin Huang}
\affiliation[inst1]{organization={Faculty of Information Technology, Monash University},%Department and Organization
            % addressline={Address One}, 
            city={Melbourne},
            postcode={3800}, 
            state={VIC},
            country={Australia}}

\author[inst2]{Hongzhi Yin}

\affiliation[inst2]{organization={Faculty of Engineering, The University of Queensland},%Department and Organization
            % addressline={Address Two}, 
            city={Brisbane},
            postcode={4072}, 
            state={QLD},
            country={Australia}}

\begin{abstract}
Knowledge tracing (KT) aims to leverage students' learning histories to estimate their mastery levels on a set of pre-defined skills, based on which the corresponding future performance can be accurately predicted. 
As an important way of providing personalized experience for online education, KT has gained increased attention in recent years. 
In practice, a student's learning history comprises answers to sets of massed questions, each known as a session, rather than merely being a sequence of independent answers. Theoretically, within and across these sessions, students' learning dynamics can be very different. 
Therefore, how to effectively model the dynamics of students' knowledge states within and across the sessions is crucial for handling the KT problem. 
Most existing KT models treat student's learning records as a single continuing sequence, without capturing the sessional shift of students' knowledge state. 
To address the above issue, we propose a novel hierarchical transformer model, named HiTSKT, comprises an interaction(-level) encoder to capture the knowledge a student acquires within a session, and a session(-level) encoder to summarise acquired knowledge across the past sessions. 
To predict an interaction in the current session, a knowledge retriever integrates the summarised past-session knowledge with the previous interactions' information into proper knowledge representations. These representations are then used to compute the student's current knowledge state. 
Additionally, to model the student's long-term forgetting behaviour across the sessions, a power-law-decay attention mechanism is designed and deployed in the session encoder, allowing it to emphasize more on the recent sessions. 
Extensive experiments on three public datasets demonstrate that HiTSKT achieves new state-of-the-art performance on all the datasets compared with six state-of-the-art KT models.
\end{abstract}


% \begin{highlights}
% \end{highlights}

\begin{keyword}
User behaviour modelling \sep Knowledge tracing \sep Educational data mining \sep Learner modeling \sep Hierarchical Transformer
\PACS 0000 \sep 1111
\MSC 0000 \sep 1111
\end{keyword}

\end{frontmatter}



\section{Introduction}
Online education breaks the temporal and spatial limitations of traditional in-person learning and brings huge educational benefits to society, especially during the Covid-19 period \cite{liu2021survey, suo2008open, emanuel2013moocs}.
It is essential to provide personalized experience for students in online education as different students have different mastering levels of specific skills or concepts during their learning progresses. 

Knowledge tracing (KT) which aims to trace students' learning histories to estimate their mastery levels on a set of pre-defined skills is essential in providing personalized experience in online education and has attracted increased attention these years \cite{song2022survey, khajah2016deep}.
Based on the traced students' knowledge, E-educators can properly understand the dynamics of students' knowledge states and provide personalized learning curricula accordingly. 
Intelligent tutoring systems (ITS), which aim to automate the task of personalized curricula recommendation, have been endowed with the same tracing ability as human educators thanks to the advancements of the knowledge tracing (KT) techniques in machine learning. More specifically, an ITS leverages such techniques to model and infer students' mastery levels of skills over time from their historical responses to exercise questions, based on which their corresponding future performance can be accurately predicted. 

% Tracing students' mastering levels of specific skills or concepts during their learning progresses plays a prominent role in education and teaching. Based on the trace, educators can properly understand the dynamics of students' knowledge states and provide personalized learning curricula accordingly. Intelligent tutoring systems (ITS), which aims to automate the task of personalized curricula recommendation, have been endowed with the same tracing ability as human educators thanks to the advancements of the knowledge tracing (KT) techniques in machine learning. More specifically, an ITS leverages such techniques to model and infer students' mastery levels of skills over time from their historical responses to exercise questions, based on which their corresponding future performance can be accurately predicted. 

Despite the notable achievements, knowledge tracing remains a challenging and unresolved issue in the field. In practice, the dynamics that underlie students' mastery of skills are characterized by complexity and evolution, yet the current research has not extensively explored these aspects.
Over the past two decades, numerous knowledge tracing models leveraging statistical and machine learning techniques have been proposed. A predominant approach involves utilizing hidden Markov models (HMMs) to capture the dynamics of students' knowledge states. These models incorporate various assumptions concerning the distributions of states and other relevant variables, such as students' response accuracy and question difficulty (e.g., as in \cite{corbett1994knowledge, d2008more, pardos2011kt}). Despite the efficacy and semantics brought to model learning, these assumptions, however, have also limited the models' applicability in reality as they are mostly too simplified to capture enough complexity of the dynamics (e.g., Gaussian prior assumptions on student expertise and question difficulty are mostly simplified treatments which ignore the actual statistics of the two). 

Recent KT models take advantage of the expressive power of deep neural networks (DNNs), such as the RNN-based \cite{piech2015deep, zhang2017dynamic, long2021tracing, liu2021hierarchical, su2021time} and the self-attention \cite{pandey2019self, ghosh2020context, pu2020deep, ren2023muloer} models, to overcome the modelling and inference limitations of the traditional models like HMMs. More specifically, RNN-based models predict each interaction based on their predecessors, where more recent ones have larger impacts on the current interaction. On the other hand, self-attention-based KT models condition the prediction for each interaction on all the other historical interactions by attending to them all at once. It has been shown by the previous research that these recent DNN-based models can bring substantial performance improvements to the knowledge tracing task in terms of students' response accuracy prediction and question recommendation \cite{piech2015deep, zhang2017dynamic, liu2019ekt, he2022multi, tan2022bidkt}.

\begin{figure}
    \centering
    \includegraphics[width=\linewidth]{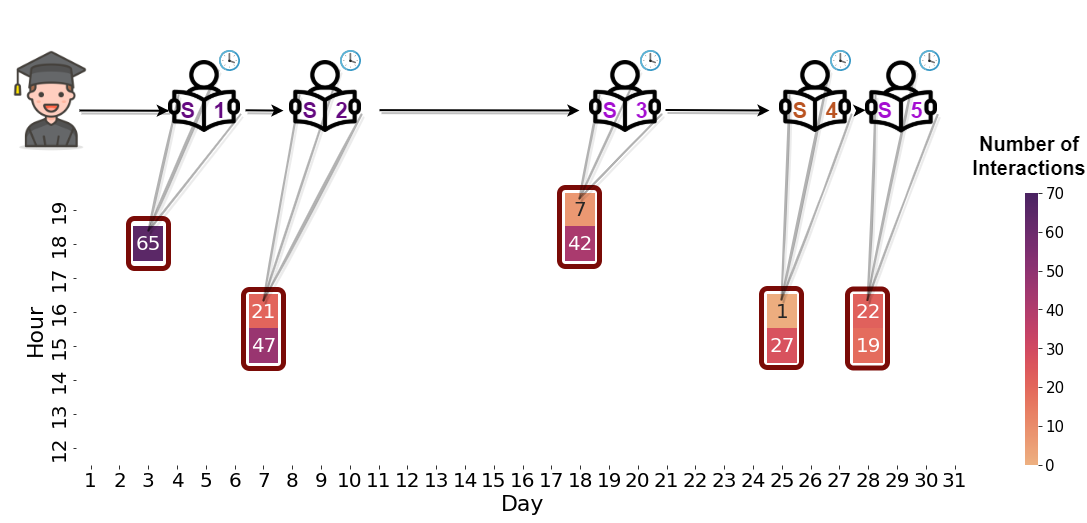}
    \caption{Interactions heatmap for one student in one month on ASSISTments2017. Each block in the red frame is a practice session containing several interactions, and time intervals are represented as arrow. One interaction means the process that a student interacts with ITS and provides one response to one exercise question. For instance, the example student has $65$ interactions at $6pm$ on the third day of that month. These $65$ interactions are relative dense and can be considered as a session practice. The following session practice happened on three days later, which contains $68$ interactions during $2$ hours on the seventh day of that month.}
    % \vspace{-1em}
    \label{17distribution}
\end{figure}

Despite the improvements brought by the DNNs, current KT models still have several known issues. One is that they model a student's historical learning records as a continuing sequence, which fails to capture how these records are distributed during the students' learning progress in reality. As an example, we show in Fig. \ref{17distribution} our observations via exploration on ASSISTments2017\footnote{ASSISTments2017 source: https://sites.google.com/site/assistmentsdata}, a real-world knowledge tracing dataset. This figure illustrates a student's learning history within 31 days, which has 5 distinct practice sessions, each containing a different number of questions answered consecutively within relatively short periods of time; while the intervals between any two consecutive practice sessions are much longer. More specifically, with a spectrum of colours that indicates the intensity levels of interactions per time unit, a session is characterized by a distinct burst of interactions with the ITS, followed by a distinct large time gap (presented as directed arrows in the figure). Such sessional behaviour of a single student repeats across the entire student population, the statistical details of which will be further introduced in Section \ref{exp_sec} to confirm the generality of our findings. Therefore, instead of being continuing, histories of students' interaction records in reality usually exhibit clustering effects by consisting of a series of separated practice sessions where each session includes a sequence of practice interactions. 

In Educational Psychology, studies of memory retention consider sessional information (e.g., in the form of massed exercises over certain spacing of time) as an indicative factor of students' learning performance\cite{brown2014science, bjork1988retrieval, rowland2015mnemonic, atkinson1968human}. According to these studies, within-session information, such as number and difficulty of the questions, and inter-session information, such as the spacing of different sessions, have different impacts on students' learning performance\cite{roediger2011critical, lyle2020amount, baddeley2012working}. Therefore, it is reasonable to separately model the within- and inter-session dynamics underlying students' learning process to capture their separate impacts. 

Existing KT models are not designed to capture the sessional information and its effects on students' performance. To do so, we propose a novel \textbf{Hi}erarchical \textbf{T}ransformer model for \textbf{S}ession-Aware \textbf{K}nowledge \textbf{T}racing (HiTSKT), that simultaneously captures the dynamics and connections underlying two types of sequences: (1) sequences of responses within a session and (2) sequences of different sessions. 

HiTSKT models these two types of sequences in a bottom-up manner with a \textit{Acquisition} \& \textit{Consolidation} (AC) modelling component. The AC component is designed as a hierarchical transformer encoder architecture, in which (1) an \textit{interaction} encoder transforms the knowledge acquired by a student within a session into an intra-session knowledge representation vector, and (2) a \textit{session} encoder receives the intra-session representations of all the past sessions, and consolidates them into a vector representing the inter-session knowledge up to the current session. The memory consolidation progress is performed by the encoder network with a forgetting mechanism that emphasizes more on the intra-session information from more recent sessions. After the inter-session knowledge gets consolidated, HiTSKT uses a \textit{Retrieval} \& \textit{Responding} (RR) modelling component to retrieve the stored inter-session knowledge representation, and integrates it with the intra-session knowledge acquired so far in the session to compute the student's current knowledge state.

The contributions of our work are summarised as follows:
\begin{itemize}
\item To our best knowledge, we are the first to exploit the underlying sessional information from the students' learning histories for the KT problem. A detailed exploratory data analysis of the sessional information on three real-world online educational datasets has been conducted in Section \ref{session_information_subsection}, which provides a new insight into the KT problem.

\item A carefully-designed hierarchical transformer-based model, named HiTSKT, is proposed to model the sessional information for the KT problem. With a session-aware hierarchical transformer encoder model and a knowledge state retrieval encoder, the model is capable of capturing students' knowledge state variations within and across the sessions, as well as their impacts on students' performance in the current session.  

\item  HiTSKT also models and captures students' sessional forgetting behaviour, which is evidenced by our exploratory data analysis in Section \ref{sec:vis}, through an power-law-decay scaled attention mechanism, designed and deployed in the acquisition \& consolidation modelling component. 

\item Extensive experiments including overall effectiveness studies, ablation studies and model interpretation have been conducted on three real-world online educational datasets. The results show that (1) the superior response correctness prediction performance of HiTSKT by capturing the sessional information, compared with six state-of-the-art KT models; (2) the effectiveness of the major components of HiTSKT. In addition, the code of HiTSKT and the implementation of several state-of-art KT models is provided\footnote{https://github.com/pokerme7777/HiTSKT}. 

\end{itemize}

The rest of the paper is organised as follows. Section \ref{relate_work_section} surveys the existing KT research related to our work. Section \ref{pro_sec} formulates the KT problem to be tackled in this paper. The details of our proposed model, HiTSKT, are described in Section \ref{methodology_sec}. Section \ref{exp_sec} concerns the experimental evaluation of HiTSKT against six state-of-the-art KT models over three real-world datasets and the discussion of the evaluation results. Section 6 draws a conclusion on the work completed in this paper, and envisage the future work to be extended from it.

\section{Related Work}\label{relate_work_section}

\begin{figure}[t]
    \centering
    \includegraphics[width=\linewidth]{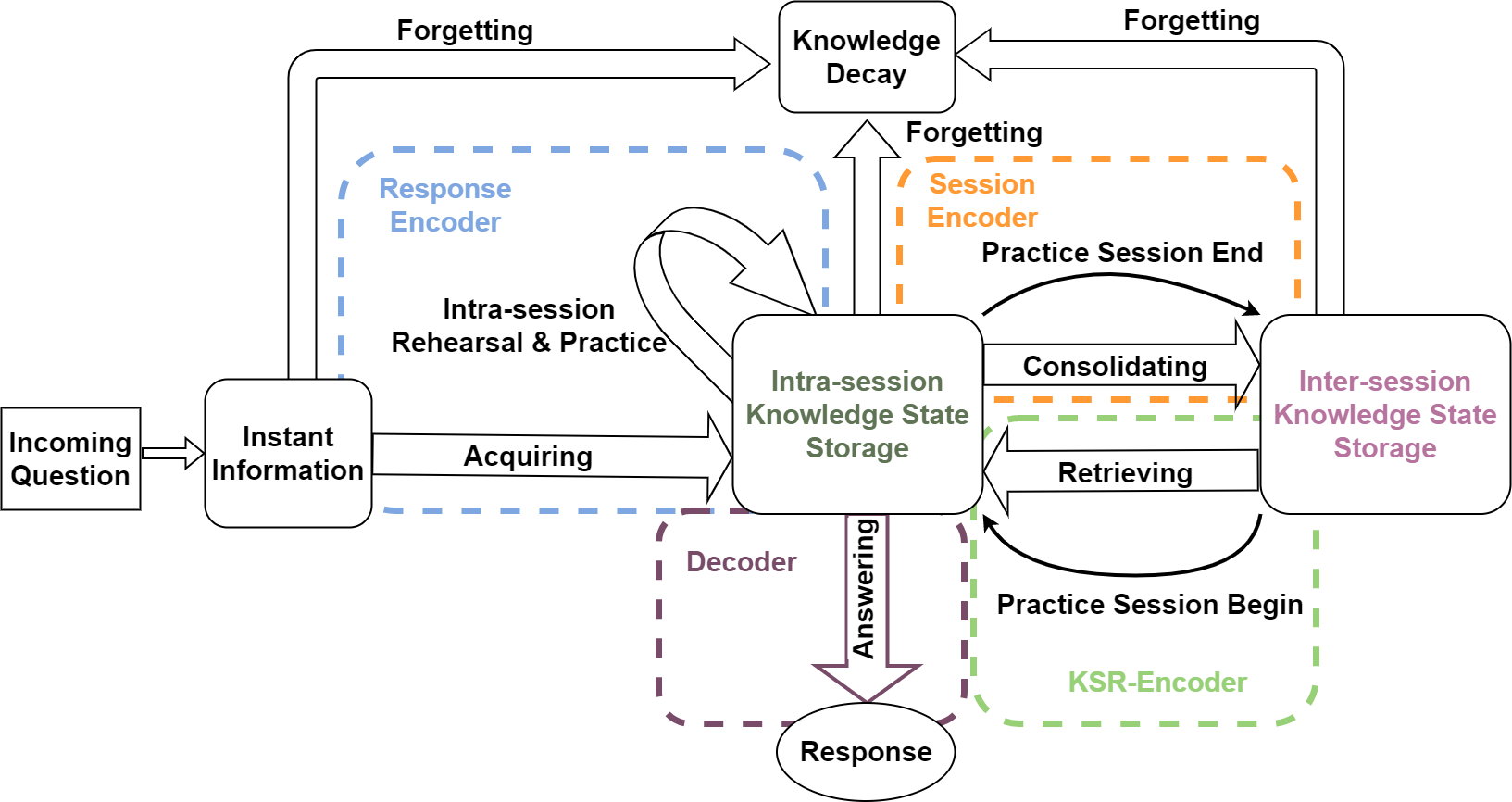}
    \caption{HiTSKT motivation based on Atkinson Shiffrin Memory Model. When students launch a session, their brain retrieves related knowledge states from their deep memory storage as the working states, which can also be viewed as intra-session knowledge states in this figure, and are ready to respond. Human intelligence acquires necessary information from the incoming question and utilizes the working states to react. Once the session practice is finished, the working states will be consolidated into the deep memory storage, which corresponds to the inter-session knowledge state, and decays over time \cite{rowland2015mnemonic, atkinson1968human, lyle2020amount}.}
    \label{as_memory_modefl}
\end{figure}

\textbf{Traditional Knowledge Tracing Methods.} Corbett and Anderson \cite{corbett1994knowledge} proposed the first Knowledge Tracing model (i.e., Bayesian Knowledge Tracing, BKT) by employing the hidden Markov model. BKT predicts students' knowledge state (i.e., their proficiency) on a skill (i.e., a knowledge concept in the curriculum) based on their performance on the immediate last quiz. They also conducted a series of experiments that confirmed the efficacy of their model. Subsequently, many extended models based on BKT have been proposed (e.g., \cite{ d2008more, pardos2011kt}) that aim to improve model performance and interpretability. However, BKT and its extended models are not capable of capturing more subtle and complex patterns underlying students' learning histories due to their intrinsic lack of flexibility \cite{khajah2016deep, abdelrahman2022deep, shen2022monitoring}. 

\textbf{Deep Learning and Other Knowledge Tracing Methods}. To address the lacks of model flexibility and prediction power encountered by the traditional KT models, Piech et al. \cite{piech2015deep} proposed the first KT model based on deep neural networks (i.e., the DKT model) with long-short term memory recurrent neural networks (LSTM). DKT outperformed BKT and its extended models on various benchmark datasets \cite{piech2015deep, gervet2020deep}. Following this pioneering work, more KT models explore the use of deep neural networks. Zhang et al. \cite{zhang2017dynamic} proposed a KT model with the dynamic key-value memory network (DKVMN) that leverages an memory-augmented neural network to extract the relationship among skills. Inspired by DKVMN, Abdelrahman and Wang \cite{abdelrahman2019knowledge} stacked a modified LSTM (i.e., the Hop LSTM) on top of a dynamic key-value network in their KT model. Furthermore, to improve model robustness and avoid model overfitting on smaller datasets, Guo et. al \cite{guo2021enhancing} proposed an adversarial training based KT method (ATKT). More recently, Wang et al. \cite{wang2021temporal} argue that students' interactions can be considered as point processes, and what happened last will have a distinct influence on learning. Hence, they proposed to use Hawkes process to model the temporal information of students' learning history and decomposes a sequence of interactions into local dynamic processes. 

A notable category of KT models is self-attentive KT models. Following the transformer's \cite{vaswani2017attention} success in natural language processing and computer vision tasks, KT models with self-attention mechanisms have been proposed. Pandey and Karypis \cite{pandey2019self} proposed the first knowledge tracing model based on self-attention mechanisms (i.e., SAKT). The SAKT model can capture the long term dependency between the interactions of one student. Pu et al. \cite{pu2020deep} proposed a KT model based on vanilla transformer \cite{vaswani2017attention}. More recently, Ghosh et al. \cite{ghosh2020context} proposed a context-aware KT model with a monotonically and exponentially decay attention mechanism to model how the student's performance from a distant past affects their current performance.

Although self-attentive KT models \cite{liu2019ekt, tan2022bidkt, guo2021enhancing, shen2020convolutional} have shown desirable tracing performance, their designs overlook sessional information and are too rigid to incorporate such information to capture the potential sessional drifts of knowledge states. %Due to the model structure limitation, these models can model the slightly fluctuated knowledge state during practice by considering histories as a single continuous sequence. But these KT models can not identify session differentiation. 
Furthermore, there have also been some models designed to capture the memory-decay dynamics underlying the variations of knowledge states caused by the students' forgetting behaviour. For example, AKT \cite{ghosh2020context} has proposed a context-aware exponentially decay attention mechanism and HMN \cite{liu2021hierarchical} uses Differentiable Neural Computer (DNC) \cite{graves2016hybrid} to improve the modeling of memory decay. However, they fail to consider the memory decay to be dependent on (the spacing of) sessions.

\textbf{Memory Retention and Educational Psychology Studies} (e.g., \cite{rowland2015mnemonic, atkinson1968human, lyle2020amount}) have shown that a student's overall performance within practice sessions can better reflect their knowledge mastery levels than their performance in a single question. They have also found that the student's knowledge states between sessions are hugely different due to the memory decay and consolidation effect of long-term memory \cite{ brown2014science, bjork1988retrieval, roediger2011critical}. On the contrary, the intra-session knowledge state is relatively steady with slight fluctuation as one single exercise can not make a significant difference in memory. Therefore, it is reasonable to learn the intra-session and inter-session knowledge states separately. Current KT models ignore the potential sessional drifts in students' memories during their learning processes. We believe that modeling such drifts can considerably account for variations of a student's knowledge states and is the key to developing more effective KT models.

% \begin{figure*}[t]
%     \centering
%     \includegraphics[width=0.7\linewidth]{AS_memory_model.png}
%     \caption{HiTSKT motivation based on Atkinson Shiffrin Memory Model. }
%     \label{as_memory_modefl}
% \end{figure*}

Furthermore, educational psychologists (e.g.,\cite{brown2014science, bjork1988retrieval, roediger2011critical}) have pointed out that the learning practice is always session based.
As illustrated in Fig. \ref{as_memory_modefl}, when students launch a session, their brain retrieves related knowledge states from their memory storage as the working states, which can also be viewed as intra-session knowledge states, and are ready to respond. Human intelligence acquires necessary information from the incoming question and utilizes the working states to react. Once the session practice is finished, the working states will be consolidated into the deep memory storage, which corresponds to the inter-session knowledge state, and decays over time \cite{rowland2015mnemonic, atkinson1968human, lyle2020amount}.

Overall, we can argue that the sessional information underlying students' learning histories should be explicitly exploited for knowledge tracing. Recently, to adapt the transformer model to more complex tasks, there has emerged research work (e.g., \cite{zhang2019hibert}) that hierarchically stacks up transformer encoders to extract individual representations from partial sequences of the entire long sequence. Inspired by these studies, in this paper, we propose a KT model with hierarchical transformer encoders to fully exploit the sessional information for better modelling the variations of students' knowledge states.

\section{Problem Formulation}\label{pro_sec}

\begin{table}[t]
\centering
\small
\caption{Notations used in this paper}
\centering
\begin{tabular}{ l l }
\hline 
 Notation & Description\\ 
 \hline 
 $I$ & The number of students \\
$K$ & The number of skills \\
$Q$ & The number of questions \\
%$F$ & The number of occurrences of the most frequent question \\[0.4ex] 
%$\gamma$ & A constant value in forgetting curve approximation function \\[0.4ex] 
%$R$ & Retrievability \\[0.4ex]
%$S$ & Stability of memory \\ [0.4ex]
%$\xi$ & Memory saving percentage \\ [0.4ex]
\hline
$i$ &  The $i$-th student or student $i$ \\
$t$ & The $t$-th time step or time $t$ \\
$ses^i_{n}$ & The $n$-th session of the student $i$ \\ 
$x$ &  An interaction $x$ \\
$q$ & A question $q$ \\
$k$ & A skill/concept $k$ \\
$f_q$ & The number of occurrences of the question $q$ \\
$a$ & The binary correctness of a response \\ 
\hline
$\boldsymbol{x}$ &  Embedding of the interaction $x$ \\ 
$\boldsymbol{d}_q$ & Embedding of the difficulty of question $q$ \\
$\boldsymbol{k}$ & Embedding of the skill $k$ \\
$\boldsymbol{f}_q$ & Embedding of the number of occurrences of question $q$ \\
$\boldsymbol{a}$ & Embedding of the correctness of a response \\
%$\boldsymbol{q}$ & A rehearsal embedding \\
%$\boldsymbol{g}$ & Embedding of personal learning ability \\[0.4ex] 
$\boldsymbol{h}^{\text{Inner}}$ & Intra-session knowledge representation vector \\ 
$\boldsymbol{h}^{\text{Inter}}$ & Inter-session knowledge representation vector \\ 
\hline
% $\boldsymbol{X}$ & Interaction matrix \\[0.4ex] 
% $\boldsymbol{Q}$ & Query matrix \\[0.4ex] 
$\boldsymbol{q}$ & Attention query vector \\
$\boldsymbol{K}$ & Key matrix \\
$\boldsymbol{V}$ & Value matrix \\
$\boldsymbol{W}$ & Mapping matrix \\
% $\pmb{\alpha}$  & Attention scores \\[0.4ex]
% $\boldsymbol{T}$  & Temperature in attention mechanism \\[0.4ex]
\hline
\end{tabular}
\label{notation_table}
\end{table}

\begin{figure}
    \centering
    \includegraphics[width=\linewidth, height = 3cm]{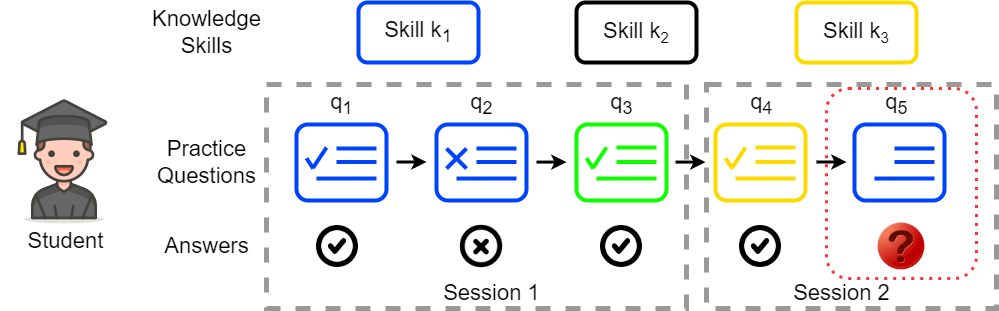}
    \caption{An illustrative example of the proposed HiTSKT Task. Given a student's past learning history including (1) first $n$ sessions i.e., $\{ses^i_1, ses^i_2,$ $\dots, ses^i_{n}\}$ and (2) the first $t-1$ interactions in the current $(n+1)$-th session, i.e., $\{x^i_{n+1,1}, x^i_{n+1,2}, \dots, x^i_{n+1,t-1}\}$, HiTSKT calculates out the student's current knowledge state first. The following step is to predict student's graded answer $a^i_{n+1,t}$ to the query question $q^i_{n+1,t}$ based on the knowledge state.}
    \label{task}
\end{figure}

Throughout the paper, the notation used to formulate our targeting KT problem are described in Table \ref{notation_table}. Based on this notation, suppose that the learning records of the student $i$ are composed of $n$ non-overlapping sessions $\{ ses^i_1, ses^i_2, \dots, ses^i_{n}\}$, where $ses^i_{n}$ denotes the student $i$'s $n$-th session.
For $ses^i_n$, it contains a sequence of $t$ interactions $\boldsymbol{x}^i_{n}=\{x^i_{n,1}, x^i_{n,2}, \dots, x^i_{n,t}\}$, where an interaction $x^i_{n,t} = \{q^i_{n,t}, k^i_{n,t}, f^i_{n,q_t} , a^i_{n,t}\}$ consists of (1) the question $q^i_{n,t}$ that the student $i$ answered at time step $t$ in session $n$, (2) the corresponding skill $k^i_{n,t}$ required for the question, (3) the number of occurrences of the question (until time $t$) $f^i_{n,q_t}$, and (4) the graded response $a^i_{n,t}$. %In other words, the interaction $\boldsymbol{x}^i_{n}$ means the student $i$ completed the question $q^i_{n,t}$ which is related to the skill $k^i_{n,t}$ with the graded answer $a^i_{n,t}$ at time step $t$ in $n$-th session. 
Note that the graded response $a^i_{n,t}\in\{1,0\}$ denotes whether the student $i$ has answered the $t$-th question in the $n$-th session correctly or not. In this case, given the student $i$'s past learning history, which includes his/her first $n$ sessions, i.e., $\{ses^i_1, ses^i_2,$ $\dots, ses^i_{n}\}$, the first $t-1$ interactions in the current $(n+1)$-th session, i.e., $\{x^i_{n+1,1}, x^i_{n+1,2}, \dots, x^i_{n+1,t-1}\}$, and a query question $q^i_{n+1,t}$ at the current time $t$, the problem of knowledge tracing concerns the prediction of the graded answer $a^i_{n+1,t}$ of student $i$ to the query question. Fig. \ref{task} illustrates our targeting KT problem formulated above where, as its distinctive feature, the differences in sessions have been highlighted with dashed frames.

In this paper, the time duration we use to define a session is $10$ hours based on human activity studies \footnote{People working hours per day \cite{nakata2011work}}. Specifically, the time interval between two sessions should be longer than the session duration. In other words, if the next interaction occurs later than $10$ hours, then this interaction belongs to a new session.

\section{Methodology}\label{methodology_sec}
HiTSKT is motivated by the Atkinson Shiffrin memory model \cite{brown2014science, bjork1988retrieval, atkinson1968human, roediger2011critical} as shown in Fig. \ref{as_memory_modefl}. 
It suggests that when students begin to practice or rehearse, their brains will first retrieve relevant knowledge from the current inter-session knowledge state storage into their intra-session knowledge state storage. The intra-session knowledge will then be integrated with the new coming information, acquired within the session thus far, as the students' current knowledge states to perform the responses. When the session ends, its final intra-session knowledge will be consolidated back (with certain extents of forgetting) into the students' inter-session storage.

\begin{figure}[ht]
  \centering
  \includegraphics[width=\linewidth]{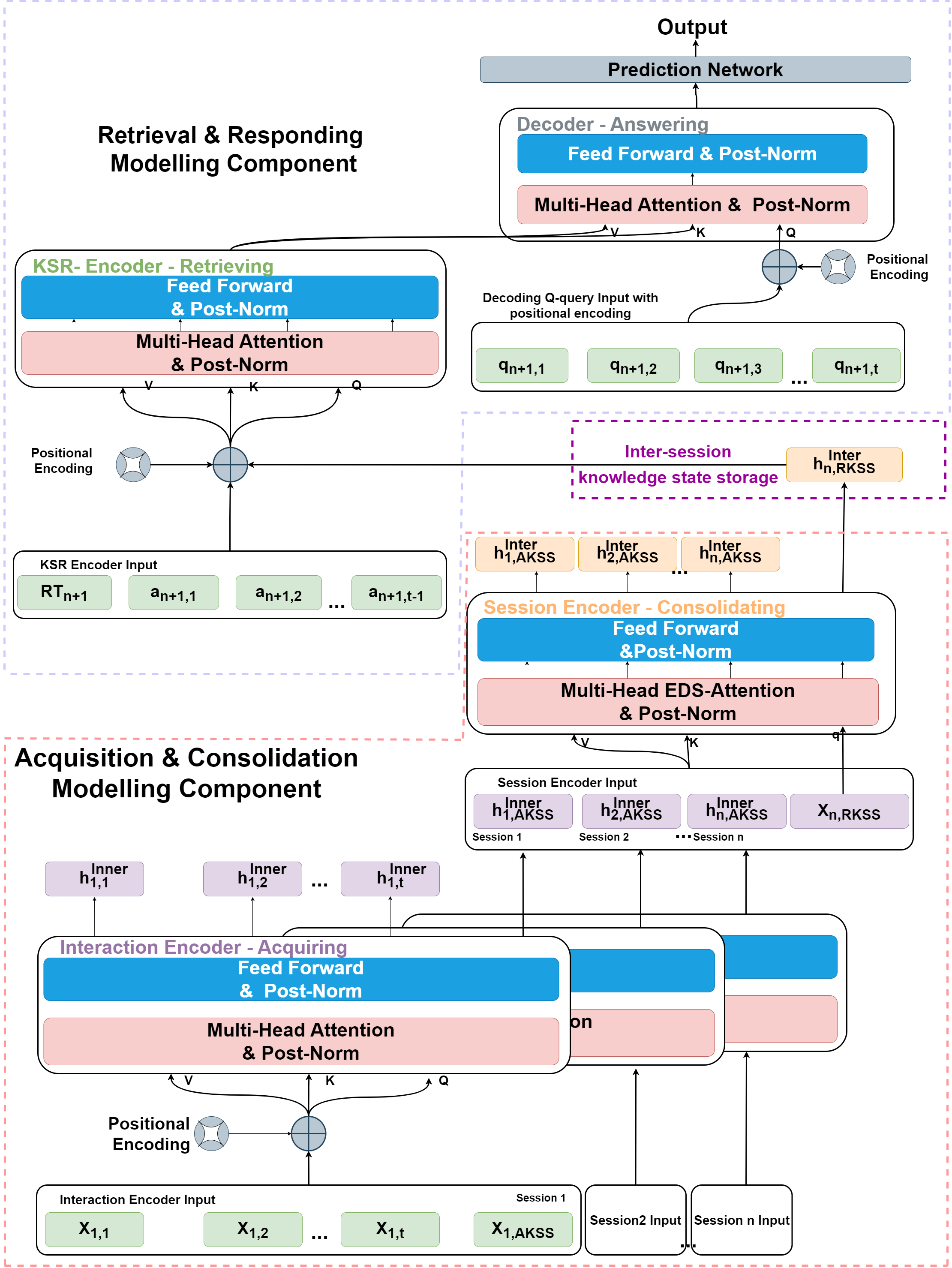}
  \caption{The overall model architecture of HiTSKT, which consists of two components: AC Component and RR Component. The AC component aims to extract and aggregate the sessional information in a bottom-up manner. Specifically, an interaction(-level) encoder first computes the knowledge representations (i.e., $\boldsymbol{h}^{\text{Inner}}_{n,\text{AKSS}}$) that summarize the acquired intra-session knowledge for each session. These representations are fed into an upper-level session encoder to summarise all the past sessions experienced by a student into an inter-session knowledge representation (i.e., $\boldsymbol{h}^{\text{Inter}}_{n,RKSS}$).The RR component leverages a KSR encoder to retrieve and refine a student's stored inter-session representation, which is then fed into an answering decoder to predict the student's next interaction in the current session. }
  \label{HiTSK_structure}
\end{figure}

Therefore, HiTSKT consists of two main components: the acquisition \& consolidation (AC) modelling component, which is a session-aware hierarchical transformer encoder model, and the retrieval \& responding (RR) modelling component with a knowledge state retrieval (KSR) encoder module and a student response prediction module. The structure of HiTSKT is shown in Fig. \ref{HiTSK_structure}. 

\subsection{Acquisition \& Consolidation Modelling Component}

The AC component aims to extract and aggregate the sessional information in a bottom-up manner. More specifically, an interaction(-level) encoder first computes the knowledge representations that summarize the acquired intra-session knowledge for each individual session from a student's learning history. Correspondingly, a special token AKSS (i.e., intrA-session Knowledge State Storage) is designed and placed alongside the input sequence to the encoder to store the intra-session knowledge representation. Then, these representations (of the AKSS tokens output from the interaction encoder) are fed into an upper-level session encoder. It proceeds to summarise all the past sessions experienced by a student into an inter-session knowledge representation, which gets consolidated into an inteR-session Knowledge State Storage (RKSS).
%This model component performs hierarchical aggregation on two-levels of interactive information: first aggregating the intra-session interactive information within each session, and then aggregating
\subsubsection{Interaction encoder}
This encoder aims to accurately capture the variations of students' knowledge states within each past session, and summarise them into the corresponding within-session knowledge representations. It comprises a rehearsal embedding layer, a multi-head attention layer and a feed forward neural network layer. 

\textbf{Rehearsal embedding} serves as the input to the interaction encoder, which should be able to capture as diverse information regarding students' individual interactions as possible. In this work, we consider such information to come from the following aspects; the skill/concept covered by a question, which can be indicative of the knowledge state correlations among questions of the same concept; the difficulty of a question, indicative of the variations of students' knowledge states, e.g. being low for difficult questions; the number of occurrences of a question, reflecting the correlations between rehearsals/practice and knowledge states; lastly, the correctness of past responses, indicative of a student's learning quality. In particular, the last two together can also be indicative of the student's knowledge state variation. For example, according to \cite{lyle2020amount, agarwal2017benefits}, it is very likely for students with high knowledge states in certain domains to correctly answer the domain questions in few times of practice. On the other hand, if a student still has low correctness on those questions after several attempts, his/her knowledge state in the particular area is more likely to be low. %It is motivated by the previous work \cite{ghosh2020context,guo2021enhancing}, which emphasizes that the KT models should avoid over-parameterization in order to address the over-fitting problem caused by the sparsity of students' historical response sequences. 

Therefore, we design the rehearsal embedding for the $t$-th interaction of the $n$-th session $\boldsymbol{x}_{n,t}$ to be the sum of the embeddings of (1) the main skill $\boldsymbol{k}_{n,t} \in \mathbb{R}^d$ of the rehearsal question $q_t$, (2) the question's difficulty $\boldsymbol{d}_{n,q_t} \in \mathbb{R}^d$, (3) the ordinal label for the occurrence number of the question $\boldsymbol{f}_{n,q_t} \in \mathbb{R}^d$ and (4) the student's response correctness $\boldsymbol{a}_{n,t} \in \mathbb{R}^d$; That is the following formulation of the rehearsal embedding $x_{n,t}$:
\begin{equation}\label{Rehearsal Embedding}
\begin{split}
 \boldsymbol{x}_{n,t} = \boldsymbol{k}_{n,t}  \oplus \boldsymbol{d}_{n,q_t} \oplus \boldsymbol{f}_{n,q_t} \oplus \boldsymbol{a}_{n,t} 
\end{split} 
\end{equation} In Section \ref{exp_sec}, we present our experimental result that the rehearsal embedding is better at modeling the KT problem, comparing with the Rasch model based embedding \cite{ghosh2020context}. Furthermore, knowledge tracing is characterized by temporal orderings of interactions. Capturing the underlying temporal dependencies between the interactions requires additional position embedding. Therefore, we apply the \textbf{positional encoding} module that uses fixed sinusoids of different frequencies \cite{vaswani2017attention} to model relative orderings, and add the corresponding position embedding to the rehearsal embedding as the final input to the interaction encoder.

\textbf{Multi-head attention layer} serves as the aggregator of the knowledge acquired within one session by performing a weighted sum over the contextualized knowledge (state) representations of each interaction into a within-session knowledge representation. To facilitate the within-session knowledge aggregation, we append a special token $\text{AKSS}$ at the end\footnote{This placement is commonly performed by the neural language models for NLP tasks.} of the input sequence to the interaction encoder. This token corresponds to and outputs the within-session knowledge representation aggregated from the contextualized representations of every interaction. In this case, the query to the attention layer is mapped from the rehearsal embedding $\boldsymbol{x}_{n, \text{AKSS}} \in \mathbb{R}^{d}$ of the special token, while the input embedding (i.e. rehearsal embedding + position embedding) of each interaction is mapped into the keys and values. More specifically, the mapping can be formulated as follows

\begin{equation}\label{Q_K_V}
\begin{split}
&\boldsymbol{q}_{n,\text{AKSS}}= \boldsymbol{W}^{T}_{Q} \boldsymbol{x}_{n, \text{AKSS}};~\boldsymbol{K}_n= \boldsymbol{W}^{T}_{K} \boldsymbol{X}_n;~\boldsymbol{V}_n= \boldsymbol{W}^{T}_{V} \boldsymbol{X}_n
\end{split}   
\end{equation}
where $\boldsymbol{W}_{Q}
, \boldsymbol{W}_{K}, \boldsymbol{W}_{V} \in \mathbb{R}^{d \times d_m}$ are learnable mapping matrices, with $d_m$ being the dimension after mapping, and $\boldsymbol{X}^{T}_n \in \mathbb{R}^{t \times d}$ is the input embedding matrix of all the interactions within the $n$-th session. Then, the scaled dot-product attention scores $\pmb{\alpha}_{n,\text{AKSS}} \in \mathbb{R}^{t}$ are computed as:

\begin{equation}\label{attention_a}
\begin{split}
&\pmb{\alpha}_{n,\text{AKSS}}= \text{Softmax}\bigg(\frac{ \boldsymbol{K}^{T}_n\boldsymbol{q}_{n,\text{AKSS}}}{\sqrt{d_{\boldsymbol{K}^{T}_n}}}\bigg)
\end{split}   
\end{equation} where $d_{\boldsymbol{K}^{T}_n}=d_m$ is the dimension of the key embedding matrix. After the Softmax function is applied to obtain $\pmb{\alpha}_{n,\text{AKSS}}$, the inner-session knowledge representation $\boldsymbol{h}^{\text{Inner}}_{n,\text{AKSS}} \in \mathbb{R}^{d_m}$ is output as:

\begin{equation}\label{attention_1}
\begin{split}
\boldsymbol{h}^{\text{Inner}}_{n,\text{AKSS}} = \sum_{t}\alpha_{n,t,\text{AKSS}}\times\boldsymbol{v}_{n,t}
\end{split}   
\end{equation}
where $\boldsymbol{v}_{n,t} \in \mathbb{R}^{d_m}$ is the value embedding corresponding to the $t$-th interaction.

\textbf{Feed-Forward neural network} (FFN) module then takes in the output $\boldsymbol{h}^{\text{Inner}}_{n,\text{AKSS}}$ from the attention layer, passes it through a two-layer multi-layer perceptron with ReLU activation to obtain the following final output of the interaction encoder: 
\begin{equation}\label{EOS_attention1}
\begin{split}
&\boldsymbol{h}^{\text{Inner}}_{n,\text{AKSS}} := \boldsymbol{\Phi}^{T}_2\sigma(\boldsymbol{\Phi}^{T}_1\boldsymbol{h}^{\text{Inner}}_{n,\text{AKSS}} + \boldsymbol{b}_1) +\boldsymbol{b}_2
\\
\end{split}   
\end{equation} where $\boldsymbol{\Phi}_1 \in \mathbb{R}^{d_m\times d'_m}, \boldsymbol{\Phi}_2 \in \mathbb{R}^{d'_m\times d_m}$ and $\boldsymbol{b}_1 \in \mathbb{R}^{d'_m}, \boldsymbol{b}_2 \in \mathbb{R}^{d_m}$ are respectively the learnable weight matrices and bias vectors of the FFN, while $\sigma(\cdot)$ denotes the ReLU activation function.

Following the above procedures, each past session in $\{ ses^i_1, ses^i_2, \dots, ses^i_{n} \}$ will be encoded by the interaction encoder into a sequence of intra-session knowledge (state) representations $\{\boldsymbol{h}^{\text{Inner}}_{1,\text{AKSS}},$ $\boldsymbol{h}^{\text{Inner}}_{2,\text{AKSS}}, \dots, \boldsymbol{h}^{\text{Inner}}_{n,\text{AKSS}}\}$. These representations will then be used to construct a historical intra-session knowledge representation matrix $\boldsymbol{H}^{T}_n\in \mathbb{R}^{n \times d}$ and be fed into the subsequent session encoder.

\subsubsection{Session Encoder}
As discussed in Section \ref{relate_work_section}, the time intervals and the intensity of the practice sessions significantly affect students' memory retention of their acquired within-session knowledge, via the memory consolidation and forgetting mechanism, thereby their long-term learning performance, which is dependent on the retrieved knowledge across sessions \cite{brown2014science,bjork1988retrieval,roediger2011critical,lyle2020amount,rowland2015mnemonic, atkinson1968human}. %Moreover, students' knowledge states are dependent on the retrieval knowledge of the inter-session knowledge states. 
Therefore, we propose to design a session-level encoder that is able to model the sessional consolidation and forgetting mechanism of students' memory by analysing their acquired within-session knowledge during their learning processes.

Based on the memory studies mentioned above, the consolidation process mainly occurs after a ``practice \& rehearsal'' session, which summarises and aggregates all the acquired knowledge from the session, and the knowledge state, to be consolidated with such knowledge, is highly dependent on the memory retention rate during the consolidation. As a result, we model the sessional knowledge ``consolidation'' by leveraging the attention mechanism, and the forgetting behaviour by explicitly modifying the attention to decay further back in time. According to the findings from the memory studies, forgetting curves are closely approximated by power-law curves \cite{donkin2012power,averell2011form,wixted1991form}. Therefore, it is reasonable to assume that the attention, cast by the memory consolidation, is subject to such forgetting curves. In other words, the influence of the past practice sessions on the consolidation will decay with a power-law effect. This leads to our design of a new monotonic, power-law-decay session-level attention mechanism.

Our proposed \textbf{memory-decay session-level attention} mechanism is based on (1) an InteR-session Knowledge State Storage (RKSS) token, and (2) a power-law-decay function on the time gaps between two sessions. More specifically, the RKSS token corresponds to the inter-session knowledge (state) representation $\boldsymbol{h}^{\text{Inter}}_{n,\text{RKSS}}$ that aggregates inner-session knowledge representations $\{\boldsymbol{h}^{\text{Inner}}_{1,\text{AKSS}},$ $\boldsymbol{h}^{\text{Inner}}_{2,\text{AKSS}}, \dots, \boldsymbol{h}^{\text{Inner}}_{n,\text{AKSS}}\}$ across all the previous sessions. The power-law-decay function that computes the decay factor $\xi_{n,j}$ from the time of the $j$-th session $T_j$ to that of the $n$-th session $T_n$ is shown as follows:  %Its representation $\boldsymbol{z}_{RKSS,\text{AKSS}}$ is a student's knowledge retention after all previous session practices:
\begin{equation}\label{forgetting_attention}
\begin{split}
&\xi_{n,j} = \frac{1}{(T_n-T_j)* S +1} \\
\end{split}   
\end{equation}
where $S$ is a tunable hyper-parameter for the stability of memory retention and $(T_n-T_j)$ is the time gap between the target session $n$ and the previous session $j$. Finally, the power-law-decay session-level attention is formulated as follows:

\begin{equation}\label{seos_attention_s}
\begin{split}
\boldsymbol{q}_{n,\text{RKSS}}= {\boldsymbol{W}^{'}_Q}^{T}\boldsymbol{x}_{n, \text{RKSS}}&;~\boldsymbol{K}^{'}_n = {\boldsymbol{W}^{'}_K}^{T}\boldsymbol{H}_n;~\boldsymbol{V}^{'}_n= {\boldsymbol{W}^{'}_V}^{T}\boldsymbol{H}_n\\
\pmb{\alpha}_{n,\text{RKSS}}&= \text{Softmax}\bigg(\frac{ {\boldsymbol{K}^{'}_n}^{T}\boldsymbol{q}_{n,\text{RKSS}}}{\sqrt{d_{{\boldsymbol{K}^{'}_n}^{T}}}}\bigg)\\
\boldsymbol{h}^{\text{Inter}}_{n,RKSS}&= \sum^{n}_{n'=1}\alpha_{n,n',\text{RKSS}} \times \xi_{n+1,n'} \times\boldsymbol{v}'_{n'} \\
\end{split}   
\end{equation} where $\boldsymbol{W}^{'}_{Q}
, \boldsymbol{W}^{'}_{K}, \boldsymbol{W}^{'}_{V} \in \mathbb{R}^{d \times d_m}$ are learnable mapping matrices of the session-level attention; $\boldsymbol{x}_{n, \text{RKSS}}$ is the rehearsal embedding of the RKSS token at the end of session $n$, which ``stores'' the inter-session knowledge before the start of session $n+1$; $\boldsymbol{v}'_{n'} \in \mathbb{R}^{d_m}$ is the $n'$-th column of the value (embedding) matrix $\boldsymbol{V}'_{n} \in \mathbb{R}^{d_m\times n}$. Finally, the acquisition \& consolidation modelling component outputs $\boldsymbol{h}^{\text{Inter}}_{n,RKSS}$, an inter-session knowledge representation for all the previous practice sessions.

\subsection{Retrieval \& Responding Modelling Component}
The RR component leverages a \textbf{knowledge state retrieval} (KSR) encoder to retrieve and refine a student's stored inter-session knowledge representation, which is then fed into an \textbf{answering decoder} module to predict the student's next interaction in the current session. The decoder computes the student's current knowledge state by integrating the inter-session representation with the intra-session information from the previous interactions in the current session.
\subsubsection{Knowledge State Retrieval Encoder}

The KSR encoder models the human memory retrieval process, in which the representation of a student's inter-session knowledge state is retrieved, and then refined by each and every previous response in the current $(n+1)$-th session, i.e., $\{a^i_{n+1,0}, a^i_{n+1,1}, a^i_{n+1,2}, \dots, a^i_{n+1,t-1}\}$, before fed into the decoder module to be used to predict the corresponding next answers $\{a^i_{n+1,1}, a^i_{n+1,2}, \dots, a^i_{n+1,t}\}$. In this case, $a^i_{n+1,0}$ is artificially padded as it does not correspond to the correctness of any question but serves as a symbol that activates the retrieval at the start of the current session. For clarity, we replace/rename it by a ``Retrieval Trigger'' (RT) token, the corresponding representation of which gets integrated with the inter-session knowledge representation to be fed into the decoder module for predicting the very first interaction $a^i_{n+1,1}$. In other words, the RT token enables the retrieval of all the necessary intra-session information preceding to the current question to be answered. In addition, the refinement of the inter-session knowledge representation $\boldsymbol{h}^{\text{Inter}}_{n,RKSS}$ at every position before the current one is its element-wise addition with the student's response correctness embedding $\boldsymbol{a}_{n,t}$ at those positions as well as their corresponding positional embeddings. Furthermore, similar to the interaction encoder, the KSR encoder pipelines a multi-head attention layer and a feed-forward neural layer to further exchange the refined knowledge at every position up to the current interaction. The final outputs of the KSR encoder serve as the key and value knowledge representations for the current query question, all of which are inputs to the decoder module.

\subsubsection{Decoder}
We leverage a transformer decoder to predict the outcome of the current interaction, where the output representations of the KSR encoder from every position up to the current one consist of the key and value embedding matrices, and the aggregated skill and difficulty embedding of the current question $q_t$ (i.e., $\boldsymbol{k}_{n+1,t} \oplus \boldsymbol{d}_{n+1,q_t}$) serves as the query embedding. Through the multi-head attention layer and the FFN layer of the transformer decoder, we obtain the final knowledge state representation for the current position (i.e. $\boldsymbol{h}^{\text{Answer}}_{n+1,t} \in \mathbb{R}^{d_m}$), which is then fed into a FFN-based prediction layer that produces the corresponding correctness prediction ($\hat{a}_{n+1,t} \in [0,1]$) for the query question. Finally, for the training of HiTSKT, the following binary cross-entropy loss between the ground-truth answers and the corresponding predicted answers over each student's individual learning histories needs to be minimised:
\begin{equation}\label{loss}
\mathcal{L}= -\sum_{i}\sum_{n}{\sum_t (a^{i}_{n,t}\log(\hat{a}^{i}_{n,t})) + ((1-a^{i}_{n,t})\log(1-\hat{a}^{i}_{n,t}))}
\end{equation}

\section{Experiments}\label{exp_sec}
In this section, we will introduce the experiments conducted to evaluate our proposed model. We train and test our model on three benchmark datasets and compare the performance of our model with state-of-the-art knowledge tracing models. Note that, the code of our model and also the implementation of the compared models are publicly available\footnote{https://github.com/pokerme7777/HiTSKT}.

\begin{table}
\centering
\small
\caption{Datasets statistics}\label{table:1}
\resizebox{0.5\textwidth}{!}{\begin{tabular}{lcccccc}
\hline
Features/Datasets  &  ASSISTments2017 & Junyi & EdNet\\
\hline
Interactions &  942,807 & 14,660,217 & 88,597,714\\
Students &  1,709 & 29,865 & 131,538\\
Questions &  3,162 & 25,630 & 13,523\\
Skills & 102 & 1,326 & 10,000\\
Avg.Sessions & 8 & 20 & 23\\
Avg.Interactions/Session & 70 & 24 & 28\\ 
Median.Interactions/Session & 54 & 16 & 16\\ 
\hline
\end{tabular}}
\end{table}

\subsection{Datasets}
Three real-world benchmark KT datasets with time stamp information (i.e., ASSISTments2017 \footnote{ASSISTments2017 source: https://sites.google.com/site/assistmentsdata}, Junyi \footnote{Junyi source: https://www.kaggle.com/junyiacademy} and Riiid's EdNet dataset \cite{choi2020ednet} ) are used to evaluate the performance of HiTSKT. The ASSISTments dataset is collected from an online high school mathematics tutoring platform. Junyi dataset consists of millions of exercise attempt logs on Junyi Academy Foundation Platform. Riiid's EdNet  data is the world's largest KT related open database containing hundreds of millions of students' interactions (records).

\textbf{Pre-processing}. As for all three datasets, we remove all the interactions where skill information is Null and drop all the interactions where the time spent is more than $9999$ seconds as these interactions might be not answered or completed. The number of occurrences of each question are also labelled by simply counting on all three datasets as well.

\textbf{Session Division}. According to the start time of the next interaction and the end time of the previous interaction information, we first calculate the time interval between each two neighbouring interactions. Then, the two consecutive interactions whose time intervals are larger than 10 hours are separated as two different sessions based on human activity studies \footnote{People working hours per day \cite{nakata2011work}}.

\textbf{Datasets Statistics.} The detailed statistics of these datasets after preprocessing are shown in Table \ref{table:1}, where Avg.Sessions is the average number of practice sessions for each student in the dataset and Avg.Interactions means number of interactions in one session on average. EdNet is the largest dataset with more than $88,000,000$ interactions, while ASSISTments2017 has the most average number of interactions ($70$) for each session. 

\begin{figure}
  \centering
  \includegraphics[width=\linewidth]{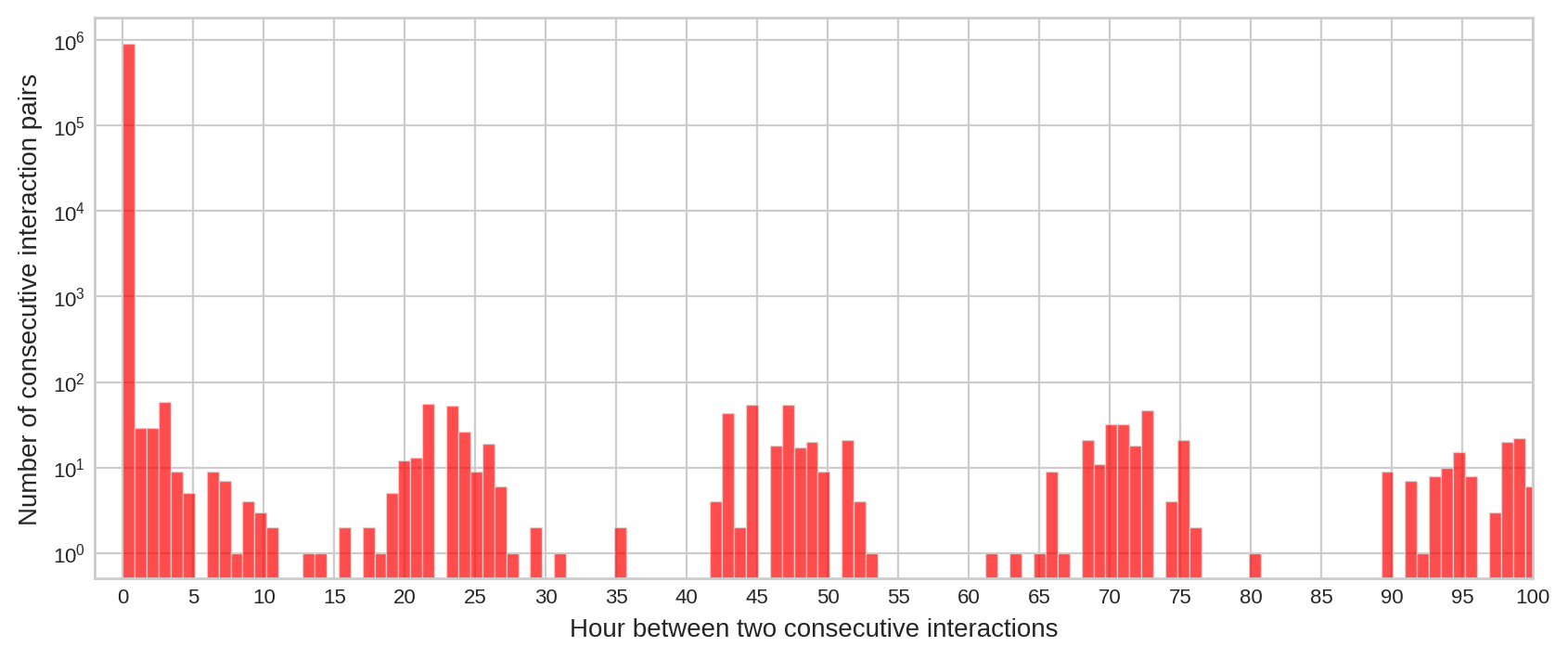}
  \caption{Histogram of time intervals between two consecutive interactions on ASSISTments2017 dataset. The x-axis means the time gap between two consecutive interactions and the y-axis means the number of interaction pairs. The time intervals between two consecutive interactions follow a long-tail distribution which means that most consecutive interactions happen within a short time interval while much fewer of them have a long time interval. 
  % Theoretically, the time intervals of inner-session interaction pairs are short. The time interval between the last interaction of one session and the first interaction of following session is long. 
  Therefore, we could conclude that histories of students' interaction records in reality usually exhibit clustering effects by consisting of a series of separated practice sessions where each session includes a sequence of practice interactions. }
  \label{time_interval}
\end{figure}

\textbf{Session Information Analysis}. \label{session_information_subsection}
To complement with Fig. \ref{17distribution} where a single student's interaction time gap distribution is shown, we conduct the statistical analysis of all the students' logs time gap distribution in Fig. \ref{time_interval} on the same dataset. In this figure, we only show the time intervals that are smaller than 100 hours as even larger intervals clearly mean different sessions. From this figure, we can see that the time intervals between two consecutive interactions follow a long-tail distribution which means that most consecutive interactions happen within a short time interval while much fewer of them have a large time interval. This together with Fig. \ref{17distribution} consolidate the universality and naturality of session information in KT task.

Compared with the other two datasets, \textbf{the information for each session in ASSISTments2017 is more abundant}.
On the other hand, the information for each student in Junyi and EdNet are sparser compared with ASSISTments2017 dataset as there are only $20$ and $23$ sessions on average for each student separately and merely $24$ and $28$ interactions in one session on average correspondingly as shown in Table \ref{table:1}. 

\begin{figure}
    \centering
    \setlength{\abovecaptionskip}{0.cm}
    \includegraphics[width=\linewidth]{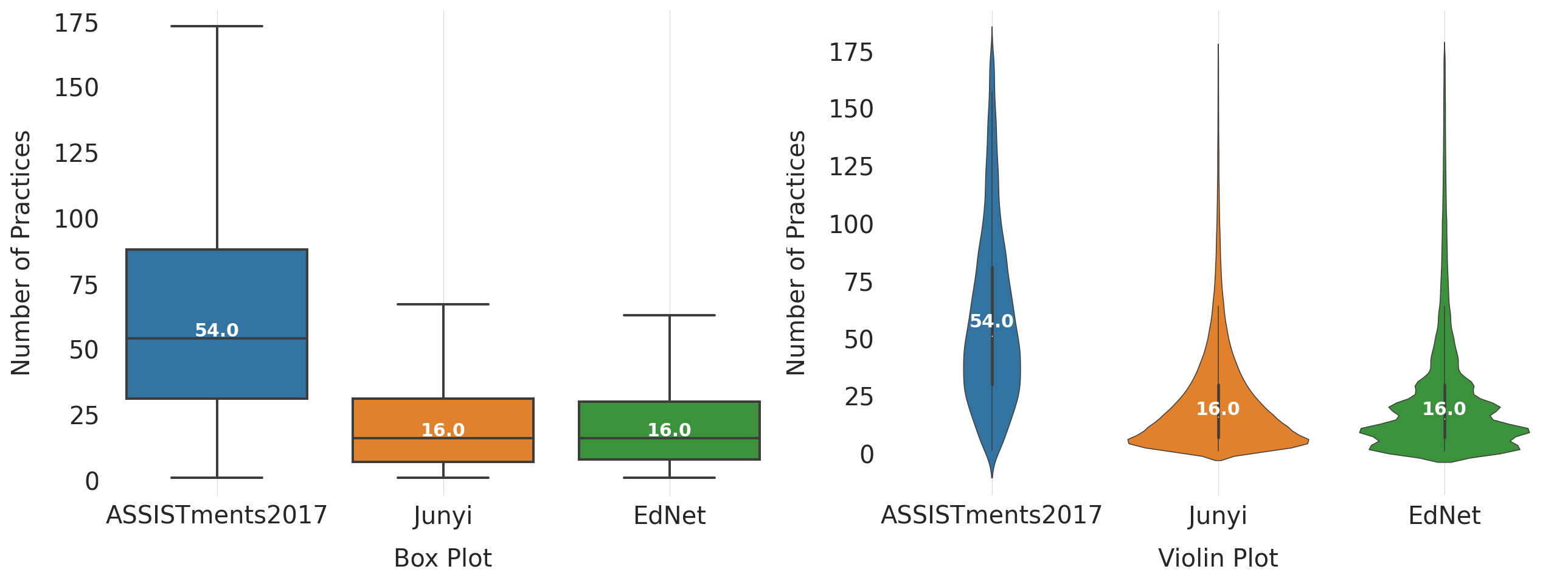}
    \caption{Box-plot and Violin-plot for the number of interactions in each session on three datasets. The distribution of the number of practice over sessions on ASSISTments2017 is close to the normal distribution, which means that the number of interactions for most sessions is around the mean number. On the contrary, the distributions are skewed to less than the mean numbers on both Junyi and EdNet datasets, which means that the session sizes are smaller and the students' histories records are sparser on these two datasets.}
    \label{boxplot}
\end{figure}

From Fig. \ref{boxplot}, the distribution of the number of practice over sessions on ASSISTments2017 is close to the normal distribution, which means that the number of interactions for most sessions are around the mean number, which is 54. On the contrary, the distributions are skewed to less than the mean numbers on both Junyi and EdNet dataset, which means that only very few sessions on Junyi and EdNet datasets have abundant practice information. Obviously, the students' histories records on Junyi and EdNet are sparser.

\textbf{Session Sequence Length and Practice Sequence Length.} Our model is transformer-based, thus the number of history sessions before the current session and the number of interactions inside each session are fixed in our model. The number of history sessions is determined by the third quartile (Q3) of all students' session numbers and we round this value to the closest power of $2$. Similarly, we choose Q3 of the number of interactions inside each session as the interaction sequence length of sessions.  The number of history sessions is 16 on ASSISTments2017, Junyi and EdNet. The number of interactions in each session is 64 on ASSISTments2017 dataset, and 32 on Junyi and EdNet datasets.

For longer interaction or session sequences, we trim the earlier interactions or sessions. For shorter sequences, padding is used to fill the vacancy.
For instance, on ASSISTments2017, $16$ is used as the number of history sessions (i.e., session sequence length) and earlier sessions are trimmed if a student has more than 16 sessions while ``session padding'' is used to fill the blank if a student has less than 16 sessions. Similarly, the most recent 64 interactions in each session is used as the interaction sequences to model the intra-session information and interaction sequences with insufficient interactions will be filled up by ``interaction padding''.  

\subsection{Baselines and Evaluation Metric}
We compare our HiTSKT model with the following advanced KT models: 
\begin{itemize}

\item DKT \cite{piech2015deep} is the first knowledge tracing model with the deep neural network. It uses long-short term memory recurrent neural network (i.e., LSTM) to make sequential prediction of students' performance. The hidden state of the LSTM can be extracted to infer the student's mastery level of different skills.
%and skills or questions information, it could be more flexible and more capable than BKT. \cite{piech2015deep}. 

\item DKVMN \cite{zhang2017dynamic} is an extended DKT model with an extra memory augmented neural network to evaluate questions and users' knowledge state separately by two matrices. Its ``key'' matrix contains the fixed representation of each skill, and its dynamic ``value'' matrix models each learner's mastery level of each skill. Also, DKVMN leverages separate ``read'' and ``write'' processes on these two matrices and be more flexible than DKT.

\item SAKT \cite{pandey2019self} is the first model utilising the self-attention mechanism to assign weights to students' responses and its structure is similar to the transformer model.

\item AKT \cite{ghosh2020context} is the state-of-the-art KT model that achieved the highest performance score on several benchmark datasets. It adopts a novel monotonic attention mechanism to model the influence of a skill learned from a distant past on a student's knowledge state. It also incorporated the difficulty coefficient to encode the skills.

\item HawkesKT \cite{wang2021temporal} is the first model that leverages Hawkes process to model the temporal information in students' learning history. More specifically, it adopts the mutual excitation mechanism and the kernel function to model the temporal cross-effect and control adaptive temporal evolution.

\item ATKT \cite{guo2021enhancing} adopts adversarial training with attentive LSTM to model the students' knowledge state. This model has a knowledge hidden state attention module that could adaptively aggregate information. Moreover, the model was trained with perturbance to improve model robustness.
\end{itemize}

The metric used to evaluate the model performance in this paper is the widely-used area under the ROC (Receiver Operating Characteristics) which is known as AUC (Area Under the Curve) \cite{zhang2017dynamic, pandey2019self, ghosh2020context, liu2019ekt, gervet2020deep, guo2021enhancing, wang2021temporal,  su2018exercise}.

\begin{table}[t]
\centering
\caption{The performance of HiTSKT and all the baselines on three datasets.}
\label{performance2}
\resizebox{0.49\textwidth}{!}{\begin{tabular}{lcccccc}
\hline
& ASSISTments2017 & Junyi & EdNet  \\
\hline
DKVMN & 0.6695 $\pm$ 0.0032  & 0.7264 $\pm$ 0.0018 & 0.5967 $\pm$ 0.0007  \\  
DKT & 0.6888 $\pm$ 0.0096 & 0.7386 $\pm$ 0.0131 & 0.6209 $\pm$ 0.0004 \\
HawkesKT &  0.6762 $\pm$ 0.0104 & 0.7296 $\pm$ 0.0124 & 0.6880 $\pm$ 0.0035 \\
SAKT &  0.7187 $\pm$ 0.0024 & 0.7852 $\pm$ 0.0014 & 0.7513 $\pm$ 0.0004\\
ATKT &  0.7417 $\pm$ 0.0030 & 0.7850 $\pm$ 0.0038 & 0.6935 $\pm$ 0.0038 \\
AKT &  0.7392 $\pm$ 0.0019 & 0.7850 $\pm$ 0.0026 & 0.7464 $\pm$ 0.0037 \\
\hline
HiTSKT & \pmb{0.7553 $\pm$ 0.0004} & \pmb{0.7911 $\pm$ 0.0001} & \pmb{0.7615 $\pm$ 0.0003} \\
\hline
\end{tabular}}
\end{table}

\begin{figure*}[t]
    \centering
    \includegraphics[width=0.8\linewidth]{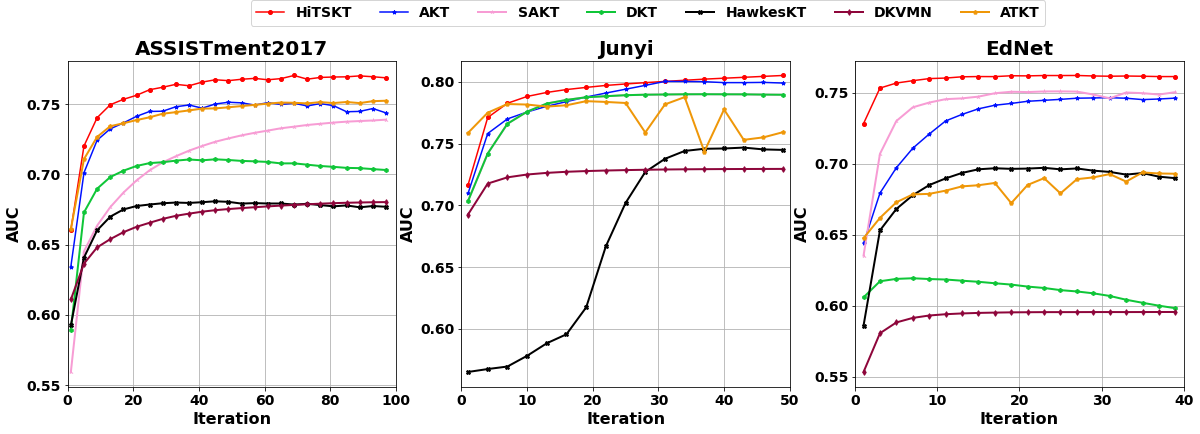}
    \caption{Convergence Curves by Validation AUC. Two self-attention baseline models (i.e., AKT, SAKT) perform better on larger datasets when compared with other baseline models. Different from the other attention based models, ATKT performs better on smaller datasets (e.g., ASSISTments2017) compared with its performance on larger datasets (e.g., EdNet). HiTSKT is robust and converges quickly than other compared models. }
    \label{epoch}
\end{figure*}

\subsection{Training and testing}
%To experiment set up, 
\textbf{Dataset Division.} We use the first $60\%$  sessions for each student to train, the next 20\% sessions to do validation and the last 20\% sessions to test. Our model does not predict the first session for all students on all datasets because our model structure needs to encode previous sessions information.

\textbf{Implementation.} HiTSKT is implemented with PyTorch, and we used Adam optimiser to train our model. All experiments are conducted on a server of which the computing core is NVIDIA V100 GPU with 16 GB memory. Constrained by the memory of the GPU, we train the model with a batch size of 64 for all datasets.

\textbf{Parameters Settings.}
As for comparison fairness, for HiTSKT and all the baselines, we tune embedding size among $\{64,128,256,$ $512\}$ and learning rate among $\{1e-5,5e-5,8e-5,1e-4,1e-3\}$ on all datasets. For parameters that are specific to one baseline, we tune them in the recommendation range presented in the paper of that baseline. 
For each model, we run it five times on all three datasets, and the average results are reported.

\begin{table*}[t]
\begin{center}
\begin{minipage}{0.8\textwidth}
\caption{HiTSKT Performance Table - Ablation Study}\label{Ablation_study}
\resizebox{\textwidth}{!}
{\begin{tabular*}{\textwidth}{@{\extracolsep{\fill}}lccc@{\extracolsep{\fill}}}
\toprule
& ASSISTments2017 & Junyi & EdNet \\
\midrule
HiTSKT using Rasch Model-Based Embeddings \cite{ghosh2020context} & 0.7417 & 0.7857 & 0.7569  \\
HiTSKT replaces KSR encoder with right-shift answer & 0.7520 & 0.7879 & 0.7588 \\
HiTSKT without AKSS & 0.7536 & 0.7897 & 0.7520 \\
HiTSKT without AKSS and RKSS & 0.7433 & 0.7774 & 0.7417 \\
HiTSKT without Positional Encoding & 0.7552 & 0.7904 & 0.7599 \\
HiTSKT using monotonic Attention & 0.7390 & 0.7841 & 0.7331 \\
\midrule
HiTSKT &  0.7553 & 0.7911 & 0.7615 \\
\toprule
\end{tabular*}}
\end{minipage}
\end{center}
\end{table*}

% \begin{figure*}
%     \centering
%     \includegraphics[width=0.8\linewidth]{epoch1.png}
%     \caption{Convergence Curves by Validation AUC. Two self-attention baseline models (i.e., AKT, SAKT) perform better on larger datasets when compared with other baseline models. Different from the other attention based models, ATKT performs better on smaller datasets (e.g., ASSISTments2017) compared with its performance on larger datasets (e.g., EdNet). HiTSKT is robust and converges quickly than other compared models. }
%     \label{epoch}
% \end{figure*}

\subsection{Main Results and Discussion}

Before we introduce the effectiveness of HiTSKT in KT problem compared with other baselines, we first have a look at the convergence rate of HiTSKT compared with other baselines. The results are shown in
the Fig. \ref{epoch} and they show that HiTSKT is robust and converges quickly than other compared models.

Table \ref{performance2} lists the performance of HiTSKT and other KT methods across three datasets and the following observations are made:
\begin{itemize}
\item HiTSKT outperforms all the compared state-of-the-art models on all three datasets (i.e., ASSISTments2017, Junyi and EdNet) which validates the effectiveness of HiTSKT in KT problem. 

\item The performance of HiTSKT is related to the session information enrichment degree. As shown in Table \ref{performance2}, compared with baselines, HiTSKT achieves more improvement on ASSISTments2017 than the improvements on Junyi and Ednet as ASSISTments2017 has richer session information as shown in Fig. \ref{boxplot}.

\item Overall, the attention based models (i.e., AKT, SAKT, ATKT and HiTSKT) perform better than the other models across all the datasets. This is consistent with the superior performance of attention based models in other areas and also confirms the correctness of designing our model based on attentive models.

\item In general, self-attention baseline models (e.g., AKT and SAKT) perform better on larger datasets when compared with other baseline models. The gaps between self-attention based models and other baseline models are minimal on the ASSISTments2017 dataset which has the least number of interactions as shown in Table \ref{table:1}. On the other hand, AKT and SAKT show more obvious improvements compared with the other baseline models on the EdNet dataset which has the most interactions as shown in Table \ref{table:1}. The underlying reason might be that these self-attention models are more complex containing more parameters that require more data to prevent the overfitting problem.

\item Different from the other attention based models, ATKT performs better on smaller datasets (e.g., ASSISTments2017) compared with its performance on larger datasets (e.g., EdNet). This might be due to the adversarial design of ATKT. The adversarial examples from the adversarial attack create notorious noise to confuse the neural network during training which makes the model more robust on small datasets. 

\end{itemize}

\subsection{Ablation Study}
\label{sec:ablation}

In this section, we present our ablation studies to validate the effectiveness of the key components of our model, including the implementation of rehearsal embedding, artificial tokens (i.e., AKSS and RKSS), Power-law Decay Attention and KSR encoder. Table \ref{Ablation_study} lists the results of ablation studies.

\textbf{Rehearsal Embedding}.
Rehearsal embedding generates the initial embedding for the inputs by mining the potential relationship between two groups of inputs including skills and questions difficulty and the student's learning ability. In this part, we compare our rehearsal embedding with the Rasch model-based embedding which is another very popular way of learning the representation for relational data in KT. Specifically, for the Rasch model-based embedding approach, we utilized the method from \cite{ghosh2020context} and IRT models \cite{khajah2014integrating}. Applying two scalars, the model measures both the questions' difficulty and the students' ability. However, the model's performance reduced by around 0.8\% on three datasets, which means rehearsal embedding is more suitable to address KT problem. Furthermore, HiTSKT's performance is still higher than others' when using Rasch model-based embedding method. That reveal the HiTSKT successfully model session-aware memory retention process.

\begin{figure}[ht]
    \centering
    \setlength{\belowcaptionskip}{0.cm}
    \includegraphics[width=\linewidth, height=4cm]{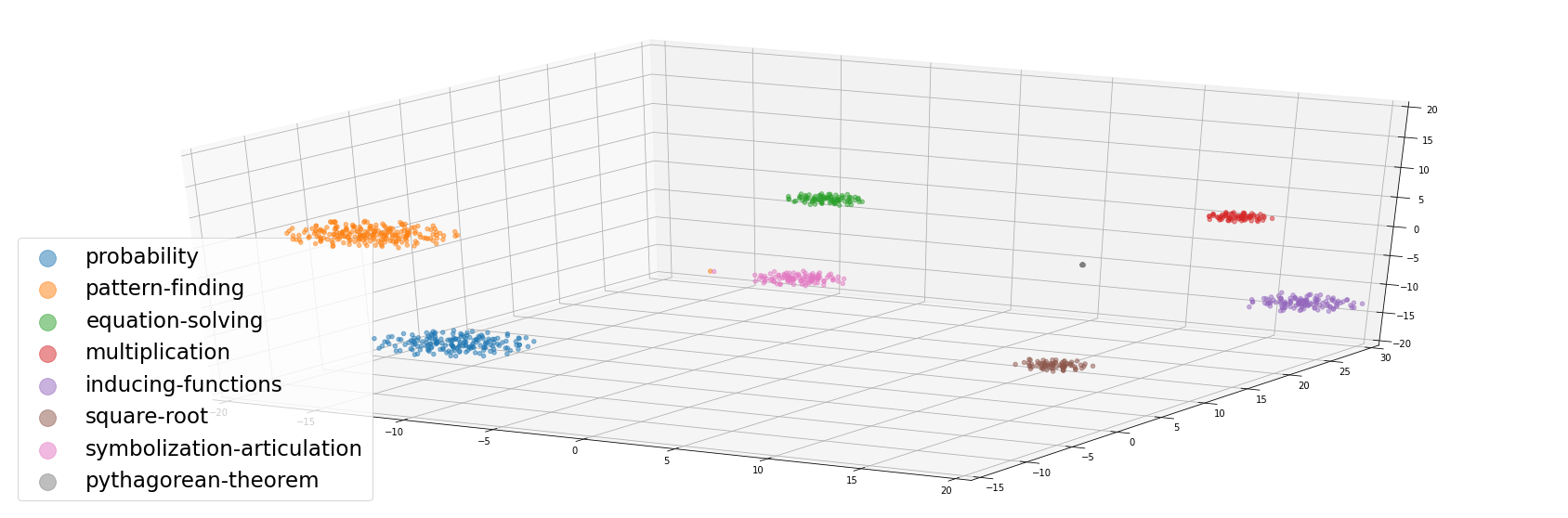}
    \caption{Question Embedding Visualisation. All questions’ difficulty are presented as dot points in three-dimensional space, and the colours represent the corresponding skills. All questions with the same skill are clustered together in space, which means questions with the same skill will be projected to close value in the embedding space. }
    \label{qkembvis1}
\end{figure}

\begin{figure*}
    \centering
    \includegraphics[width=\linewidth]{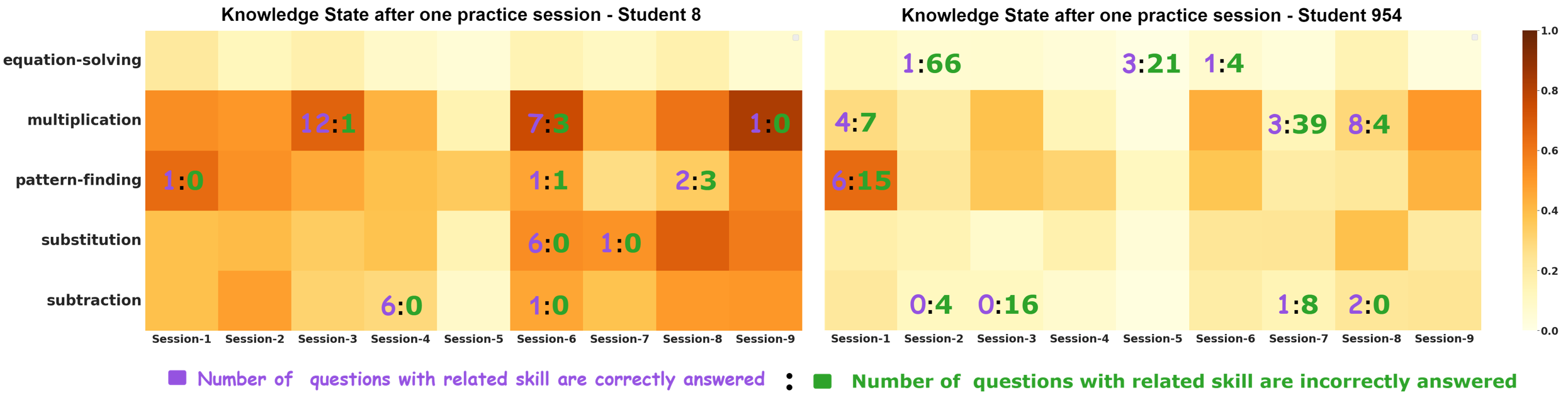}
    \setlength{\abovecaptionskip}{0.cm}
    \caption{Knowledge State Visualisation. This is an exploration figure about two students' predicted knowledge state on five skills (e.g., multiplication, equation-solving, pattern-finding, subtraction and substitution). Each block colour represents the predicted knowledge state (the weighted arithmetic mean of students' probabilities of correctly answering related questions) of corresponding skills after a practice session. The numbers represent the number of questions with related skills that are correctly/incorrectly answered in reality.}
    \label{KSV}
\end{figure*}

\begin{figure}
    \centering
    \includegraphics[width=0.95\linewidth]{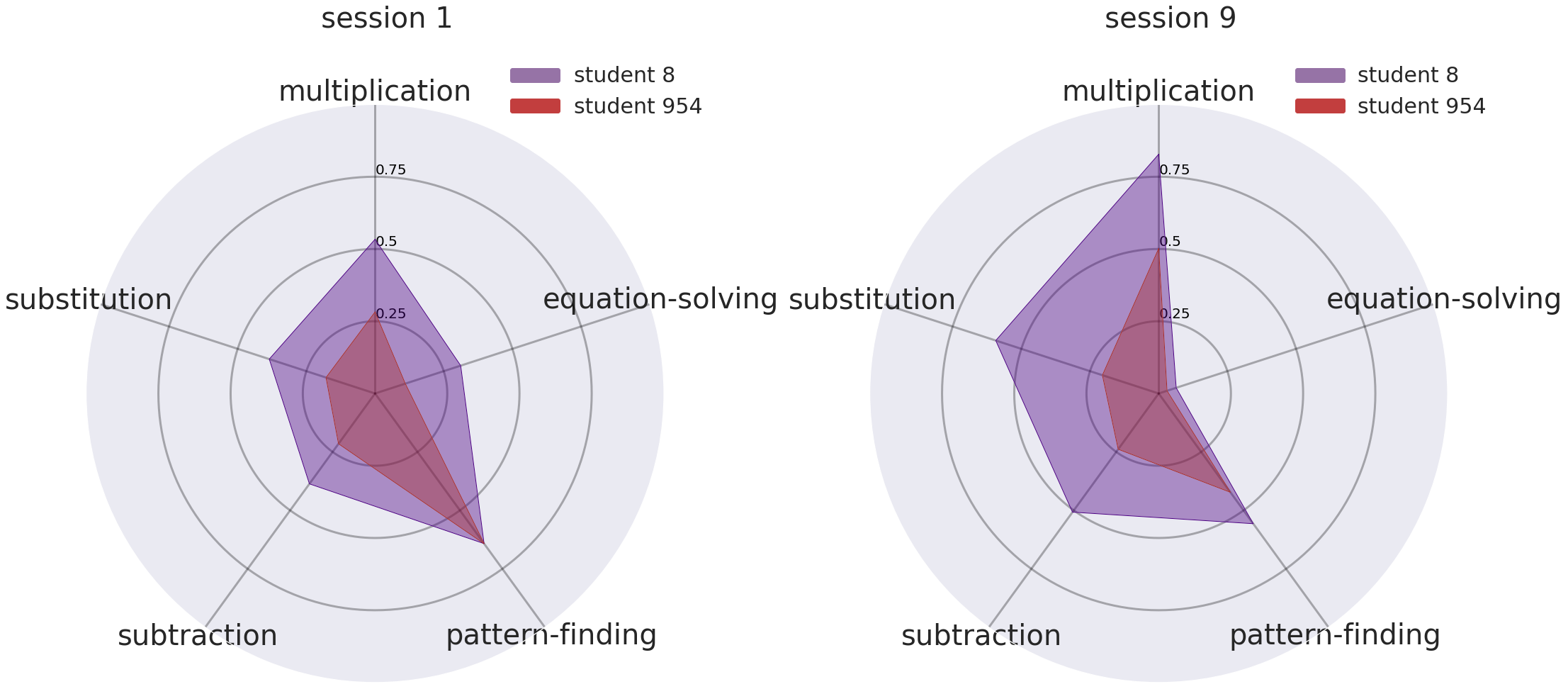}
    \caption{Radar plots for example students' Knowledge State on five skills. Two demonstrated students' skills have been improved or forgotten after nine practice sessions and students' learning rates are different.}
    \label{sig_radar}
\end{figure}

\textbf{AKSS and RKSS}. AKSS and RKSS are designed to model the hierarchical structure. The underlying technical challenge in modelling this hierarchical structure is that we need to condense a sequence of interaction representations matrix into one vector for each session. Apart from AKSS and RKSS, Pooling is a standard method to do this. In this experiment, we use Average-Pooling (Eq.\ref{avg-pooling}) to replace the AKSS and RKSS tokens. The $\kappa$ is the kernel size and the stride is the size of the window.

\begin{equation}\label{avg-pooling}
\begin{split}
\text{out}(d_1,d_2) = \frac{1}{\kappa}\sum_{m=0}^{\kappa-1}\text{input}(\text{stride} \times d_1 + m, d_2)
\end{split}   
\end{equation}

The AUC of the model without AKSS reduced by around 0.3\% compared with HiTSKT, and a further 0.6\% fall can be observed if HiTSKT does not have both artificial tokens on three datasets. Theoretically, average pooling can capture information from all the new representations of $\boldsymbol{h}^{\text{Inner}}_{n,t}$ or all the session histories representations $\boldsymbol{z}_{n,t}$. However, according to our experiment results, artificial tokens are more robust and significantly improve model performance.

\textbf{Power-law Decay Attention}. We designed the Power-law Decay Attention mechanism to capture students' forgetting behaviour over time. In this part, we compare this attention mechanism with standard scaled dot-product attention. The performance dropped to 73.90\%, 78.41\% and 73.31\% respectively on ASSISTments2017, Junyi and EdNet after we replaced the power-law decay attention with the standard one.

\textbf{KSR encoder}. When we replaced the KSR encoder with right shift answer embedding, the models' performance dropped to 75.20\%, 78.79\% and 75.88\% on ASSISTments2017, Junyi and EdNet respectively.

\subsection{Further Model interpretation through Visualisation}
\label{sec:vis}

In this session, we will interpret HiTSKT by visualising questions' difficulty $d_{n,t}$ embedding and exploring example students' knowledge state using ASSISTments2017.

\textbf{Visualising Questions' Difficulty Embedding}. We first select eight high-frequency skills, as shown in Fig. \ref{qkembvis1} legend, and get all corresponding questions' difficulty embedding in ASSISTments2017.
For these questions' difficulty, we can get their 256-dimensional embeddings after training the model. Then t-SNE \cite{van2008visualizing} is used to reduce the dimension to 3 to better visualise them.
%Then we apply Question Embedding to project these questions onto $256$ dimension embedding space and reduce the dimension to $3$ by using t-SNE \cite{van2008visualizing}.
In Fig. \ref{qkembvis1}, all questions' difficulty are presented as dot points in three-dimensional space, and the colours represent the corresponding skills. We observe that all questions with the same skill are clustered together in space, which means questions with the same skill will be projected to close value in the embedding space.  Thus, predicted students' performance on the questions will highly relate to skill mastery, which is consistent with cognitive. In other words, HiTSKT reasonably projects questions' difficulty into embedding space which will enable a more effective calculation of students' knowledge state.

\textbf{Exploration on Knowledge State Predicted by HiTSKT}. This part is to show that HiTSKT could effectively predict students' \textbf{knowledge state} after each practice \textbf{session} and the prediction is interpretable.

We select two example students (i.e., student 8 and student 954) and explore their predicted knowledge state on five skills (e.g., multiplication, equation-solving, pattern-finding, subtraction and substitution) as a demonstration. For the two students, we calculate the weighted arithmetic mean of their probabilities of correctly answering related questions for each skill after each practice session.

These mean probabilities are used as the predicted students' knowledge state (i.e., mastery level) on the corresponding skills after each practice session as shown in Fig. \ref{KSV}.
Each block represents the predicted knowledge state of corresponding skills after a practice session. For instance, the top left block in the first subplot means the predicted knowledge state on skill (equation-solving) after session-$1$ is close to 0.3. 
The numbers in Fig. \ref{KSV} represent the number of questions with related skills that are correctly/incorrectly answered.
Based on this, we can tell example student 8 is an excellent student with 0.7656 overall accuracy on these five skills from session $1$ to $10$, while example student 954 only has 0.1776 overall accuracy.
Based on Fig. \ref{KSV}, two observations can be made. 

First, forgetting and memory re-consolidation are existing phenomena in knowledge tracing and HiTSKT is able to model these behaviours. For example, student 8 did many practices in terms of multiplication in session 3 and the accuracy rate was $12/(12+1)$. If there is no forgetting, then she/he should be able to well answer questions related to multiplication in all the later sessions. However, we can see that she/he showed worse performance (i.e., $7/(7+3)$ compared with $12/(12+1)$) in session 6. This validates the existence of forgetting behaviour. Later, as more interactions were done related to this skill, student 8's master level in this skill got re-consolidated in session 9. Our model is able to capture these forgetting and re-consolidation behaviours as we can see that the colour got lighter and lighter after session 3  and the colour got darker and darker after session 6 as time went by. 

Second, our model is able to effectively capture students' knowledge state change as practice goes on. This actually echoes with our first observation as it shows that our model is able to capture how student 8's knowledge state changes as she/he does more practice. Another support is the example of student 954 learning equation-solving. As this student showed bad performance in all the interactions related to this skill, our model always predicts a bad master level of this student in terms of this skill.

Overall, we know two students' skills have been improved after nine practice sessions and students' learning rates are different from Fig. \ref{sig_radar}. Student 8 made more remarkable progress, although only half the quantity questions of student 945 did during this period. Effective practice sessions can boost student knowledge (i.e., Multiplication of student 8), while recommending practice questions casually will waste students' time and have no benefit to students (i.e., Equation-solving of student 954). Meanwhile, the recommendation system should be cautious about forgetting phenomena and help students revisit a suitable time stamp (i.e., Equation-solving of student 8).

According to the visualisation and analysis above, we can tell that the prediction from HiTSKT is highly consistent with the students' actual situation and fits the memory law mentioned.

\section{Conclusion}
In this paper, we focus on exploiting the session information from students' learning histories for tackling the knowledge tracing problem, as inspired by the detailed exploratory data analysis of session information over several real-world knowledge tracing datasets. To do so, we have introduced a carefully designed hierarchical transformer model for session-aware knowledge tracing. Our proposed model, HiTSKT, consists of two main components, which are the \textit{acquisition} \& \textit{consolidation} modelling component with a hierarchical transformer encoder architecture to summarise two types of acquired knowledge: the (lower-level) intra-session and (higher-level) inter-session knowledge, and the \textit{retrieval} \& \textit{responding} modelling component with a knowledge retrieval module (i.e. the KSR encoder) and an answering (transformer) decoder module. Furthermore, we have designed a power-law-decay attention mechanism for HiTSKT that captures students' forgetting behaviour over their acquired inter-session knowledge as a result of long-term sessional drifts.

Extensive experiments have been conducted on three large-scale real-world knowledge tracing datasets and the results show that HiTSKT achieves new state-of-the-art performance on all the datasets in terms of future response correctness prediction. Furthermore, our ablation studies have also validated the effectiveness of all the key components of HiTSKT. Visualisation has also been provided which shows that HiTSKT is interpretable and is able to learn students' knowledge states effectively. Directions of future work include (1) designing more sophisticated memory-decay attention mechanism based on state-of-the-art memory retention and decay studies in Educational Psychology, and (2) investigating whether masking language modeling, a powerful technique of training transformer-based deep language models in NLP, can be adapted and applied to enable more effective training of HiTSKT for knowledge tracing.

\appendix

% \section{Sample Appendix Section}
% \label{sec:sample:appendix}
% Lorem ipsum dolor sit amet, consectetur adipiscing elit, sed do eiusmod tempor section \ref{sec:sample1} incididunt ut labore et dolore magna aliqua. Ut enim ad minim veniam, quis nostrud exercitation ullamco laboris nisi ut aliquip ex ea commodo consequat. Duis aute irure dolor in reprehenderit in voluptate velit esse cillum dolore eu fugiat nulla pariatur. Excepteur sint occaecat cupidatat non proident, sunt in culpa qui officia deserunt mollit anim id est laborum.

 \bibliographystyle{elsarticle-num} 
 \bibliography{cas-refs}





% \end{thebibliography}
\end{document}